\documentclass[conference]{IEEEtran}
\IEEEoverridecommandlockouts
\usepackage{cite}
\usepackage{amsmath,amssymb,amsfonts}
\usepackage{algorithmic}
\usepackage{graphicx}
\usepackage{textcomp}
\usepackage{xcolor}
\usepackage{subfig}
\usepackage{multicol}

\def\BibTeX{{\rm B\kern-.05em{\sc i\kern-.025em b}\kern-.08em
    T\kern-.1667em\lower.7ex\hbox{E}\kern-.125emX}}
\begin{document}

\title{On the Reduction of Variance and Overestimation of Deep Q-Learning\\}

\author{\IEEEauthorblockN{Mohammed Sabry}
\IEEEauthorblockA{\textit{University of Khartoum, Faculty of Engineering}\\
\textit{Department of Electrical and Electronic Engineering}\\
Khartoum, Sudan \\
mhmd.sabry.ab@gmail.com}
\and
\IEEEauthorblockN{Amr M. A. Khalifa}
\IEEEauthorblockA{\textit{African Institute for Mathematical Sciences} \\
Kigali, Rwanda\\
amr.khalifa.m.a@gmail.com}
}
\maketitle

\begin{abstract}
The breakthrough of deep Q-Learning on different types of environments revolutionized the algorithmic design of Reinforcement Learning to introduce more stable and robust algorithms, to that end many extensions to deep Q-Learning algorithm have been proposed to reduce the variance of the target values and the overestimation phenomena. In this paper, we examine new methodology to solve these issues, we propose using Dropout techniques on deep Q-Learning algorithm as a way to reduce variance and overestimation. We also present experiments conducted on benchmark environments, demonstrating the effectiveness of our methodology in enhancing stability and reducing both variance and overestimation in model performance.
\end{abstract}


\section{Introduction}
Reinforcement Learning (RL) is a learning paradigm that solves the problem of learning through interaction with environments, this is a totally different approach from the other learning paradigms that have been studied in the field of Machine Learning namely the supervised learning and the unsupervised learning. Reinforcement Learning is concerned with finding a sequence of actions an agent can follow that could lead to solve the task on the environment \cite{b1}\cite{b2}\cite{b3}. Most of Reinforcement Learning techniques estimate the consequences of actions in order to find an optimal policy in the form of sequence of actions that can be followed by the agent to solve the task. The process of choosing the optimal policy is based on selecting actions that maximize the future payoff of an action. Finding an optimal policy is the main concern of Reinforcement Learning for that reason many algorithms have been introduced over a course of time, e.g, Q-learning\cite{b4}, SARSA\cite{b5}, and policy gradient methods\cite{b6}. These methods use linear function approximation techniques to estimate action value, where convergence is guaranteed \cite{b7}. However, as challenges in modeling complex patterns increase, the need for expressive and flexible non-linear function approximators becomes clear. The recent advances in deep neural networks helped to develop artificial agent named deep Q-network(DQN)\cite{b8} that can learn successful policies directly from high-dimensional features. Despite the remarkable flexibility and the huge representative capability of DQN, some issues emerge from the combination of Q-learning and neural networks. One of these issues, known as "overestimation phenomenon," was first explored by \cite{b9}. They noted that the expansion of the action space in the Q-learning algorithm, along with generalization errors in neural networks, often results in an overestimation and increased variance of state-action values. They suggested that to counter these issues, further modifications and enhancements to the standard algorithm would be necessary to boost training stability and diminish overestimation. In response, \cite{b10} introduced Double-DQN, an improvement that incorporates the double Q-learning estimator \cite{b11}, aiming to address the challenges of variance and overestimation. Additionally, \cite{b31} developed the Averaged-DQN algorithm, a significant improvement over the standard DQN. By averaging previously learned Q-values, Averaged-DQN effectively lowers the variance in target value estimates, thus enhancing training stability and overall performance.

In this paper, we introduce and conduct an empirical analysis of an alternative approach to mitigate variance and overestimation phenomena using Dropout techniques. Our main contribution is an extension to the DQN algorithm that incorporates Dropout methods to stabilize training and enhance performance. The effectiveness of our solution is demonstrated through computer simulations in a classic control environment. 

\section{Background}
\subsection{Dropout}
Deep neural networks are the state of the art learning models used in artificial intelligence. The large number of parameters in neural networks make them very good at modelling and approximating any arbitrary function. However the larger number of parameters also make them  particularly prone to over-fitting, requiring regularization methods to combat this problem. Dropout was first introduced in 2012 as a regularization technique to avoid over-fitting\cite{b12}, and was applied in the winning submission for the Large Scale Visual Recognition Challenge that revolutionized deep learning research\cite{b13}. Over course of time a wide range of Dropout techniques inspired by the original method have been proposed. The term Dropout methods was used to refer to them in general\cite{b14}. They include variational Dropout\cite{b15}, Max-pooling Dropout\cite{b16}, fast Dropout\cite{b17}, Cutout\cite{b18}, Monte Carlo Dropout\cite{b19}, Concrete Dropout\cite{b20} and many others.
\begin{figure}[htbp]
\centerline{\includegraphics[width=.5\textwidth]{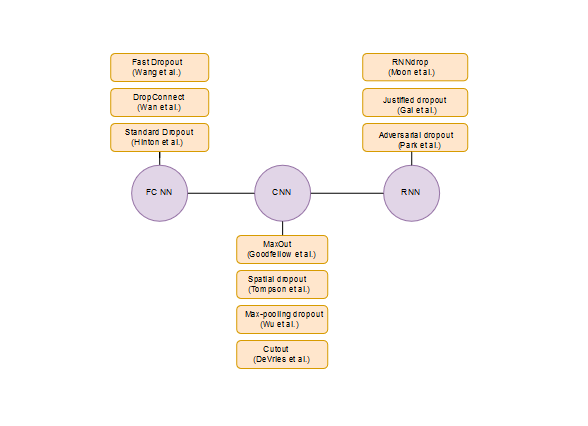}}
\caption{Some proposed Dropout methods for different neural networks architectures.}
\label{fig1}
\end{figure}
\subsubsection{Standard Dropout}
It's the original Dropout method. It was introduced in 2012. Standard Dropout provides a simple technique for avoiding over-fitting in fully connected neural networks\cite{b12}. During each training phase, each neuron is excluded from the network with a probability \textit{p}. Once trained, in the testing phase the full  network is used, but each of the neurons' output is multiplied by the probability \textit{p} that the neuron was excluded with. This approach gives approximately the same result as averaging of the outcome of a great number of different networks which is very expensive approach to evaluate, this compensates that in the testing phase Dropout achieves a green model averaging. The probability can vary for each layer, the original paper recommend $\textit{p} = 0.2$ for the input layer and $\textit{p }= 0.5$ for hidden layers. Neurons in the output layer are not dropped. This method proved effective for regularizing neural networks, enabling them to be trained for longer periods without over-fitting and resulting in improved performance, and since then many Dropout techniques have been improved for different types neural networks architectures (Figure \ref{fig1}).

\subsection{Reinforcement Learning}
\begin{figure}[htbp]
\centerline{\includegraphics[width=.5\textwidth]{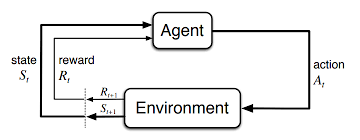}}
\caption{Markov Decision Process (Famous RL framework).}
\label{fig2}
\end{figure}

The general framework of Reinforcement Learning \cite{b21} (Figure \ref{fig2}) describes an agent engaged in a sequential decision-making problem via its interactions with an environment, occurring at discrete time steps ($t = 0, 1, \dots$). At each time step $t$, the agent observes a state $s_t \in S$, chooses an action $a_t \in A$, and receives a reward $r_t \in R$. This choice leads to a transition to the next state $s_{t+1} \in S$. The objective function considered is the discounted cumulative reward, defined as: $ G_t = \sum_{k=0}^{\infty}\gamma^k R_{t+k+1}$ where $\gamma \in [0, 1]$ is the discount factor. 
The agent's goal is to determine an optimal policy $\pi : S\rightarrow A$ that maximize the expected discounted cumulative reward. Reinforcement learning methods develop policies by utilizing value functions, which assess the cumulative reward value of a state or state-action pair to establish a policy $\pi$ that maximizes the expected discounted cumulative reward from a specified state $s$.
Specifically, our focus is on state-action value functions, which are defined as follows:
\begin{equation}
Q^\pi(s,a) = E^\pi[ \sum_{t=0}^{\infty}G_t |s_0 = s, a_0 = a]
\end{equation}

The optimal Value function denoted by:
\begin{equation}
Q^*(s,a) = max_\pi Q^\pi(s,a)
\end{equation}
The concept of the optimal value function can be extended to other scenarios, including those with continuous time steps, continuous action spaces, and continuous state spaces.    
\subsubsection{Q-Learning}
Q-learning is among the most widely used reinforcement learning (RL) algorithms\cite{b4}. It's based on an incremental dynamic programming technique because of the step by step look-up table representation in which it determines the optimal policy\cite{b22}. The Q-learning algorithm employs a table to estimate the optimal action value function, $Q^*$. This table encompasses all states and actions within the environment and utilizes the value function to assess the quality (Q-function) of state-action pairs. It then updates using the following rule:
\begin{equation}
\begin{split}
Q(s,a) \leftarrow & \; Q(s_t,a_t) \\
                  & + \alpha \left( r + \gamma \max_{a_{t+1}} Q(s_{t+1},a_{t+1}) - Q(s,a) \right)
\end{split}
\end{equation}
where $s_{t+1}$ is the resulting state after applying action \textit{a} in the state \textit{s}, \textit{r} is the immediate reward observed for action \textit{a} at state \textit{s}, $\gamma$ is the discount factor, and $\alpha$ is learning rate. 

The use of a look-up table for representing state-action values encounters limitations when dealing with a large number of states and actions. Maintaining a table that includes all possible state-action pair values in memory is impractical. A common approach to address this challenge is to adopt alternative representations, such as those provided by Neural Networks and Deep Neural Networks.
\subsubsection{Deep Q-learning (DQN)}
Deep Q-learning combines Q-learning algorithm with neural network approximation to approximate the action-value function $Q(s, a,\theta)$\cite{b8}.
Then the update rule of Q values become:
\begin{equation}
L_i(\theta_i) =  (r+\gamma max_{a'} Q(s',a';\theta_i^-) - Q(s,a;\theta_i))^2
\end{equation}
$\theta_i$ and $\theta_i^-$ are the parameters of network and target network at iteration \textit{i} respectively. The target network parameters $\theta_i^-$ are only updated with the Q-network parameters $\theta_i$ every \textit{C} steps and are held fixed between individual updates. DQN uses a memory bank approach to store(s,a,r,s') as experiences from previous iterations sampled uniformly and used in next iterations this technique termed as Experience Reply(ER).

\subsubsection{Overestimation phenomenon}
This phenomenon introduces a positive bias that may lead to asymptotically sub-optimal policies, distorting the cumulative rewards. The majority of analytical and empirical studies suggest that overestimation typically stems from the max operator used in the Q-learning value function. Additionally, the noise from approximation methods and, in some cases, environmental factors can exacerbate the expected overestimation. This often results in a more pronounced bias in states where Q-values for different actions are similar, or in states that mark the beginning of a long trajectory.

\subsubsection{DQN Variance}
The sources of DQN variance are Approximation Gradient Error(AGE)\cite{23} and Target Approximation Error(TAE)\cite{24}. In Approximation Gradient Error, the error in gradient direction estimation of the cost function leads to inaccurate  and extremely different predictions on the learning trajectory through different episodes because of the unseen state transitions and the finite size of experience reply buffer. This type of variance leads to converging to sub-optimal policies and brutally hurts DQN performance. The second source of variance Target Approximation Error which is the error coming from the inexact minimization of DQN parameters. Many of the proposed extensions focus on minimizing the variance that comes from AGE by finding methods to optimize the learning trajectory  or from TAE by using methods like averaging to exact DQN parameters. Dropout methods have the ability to assemble these two solutions which minimize different source of variance. Dropout methods can achieve a consistence learning trajectory and exact DQN parameters with averaging, which comes inherently with Dropout methods.\\
In the experiments we detected variance using standard deviation from average score collected from many independent learning trails.
\begin{figure}[htbp]
\centerline{\includegraphics[width=.5\textwidth]{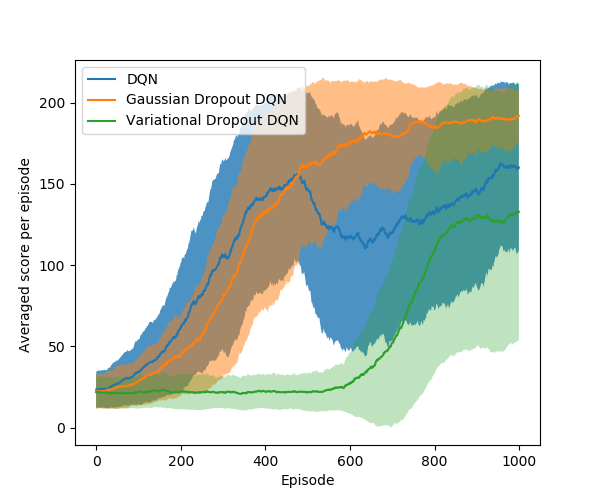}}
\caption{Dropout DQN with different Dropout methods in CARTPOLE environment. The bold lines represent the average scores obtained over 10 independent learning trials, while the shaded areas indicate the range of the standard deviation.}
\label{fig3}
\end{figure}
\\
\\
\begin{figure}[htbp]
\centerline{\includegraphics[width=.5\textwidth]{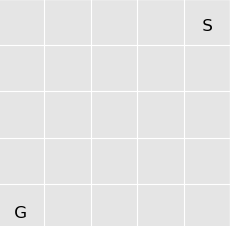}}
\caption{Gridworld problem. The agent starts its journey from the upper-right corner of the grid, while a reward of +1 is positioned in the bottom-left corner.}
\label{fig4}
\end{figure}

\section{Experiments}

Our experiments were organized to explore the following questions:

\begin{itemize}
\item How does Dropout influence the variance in DQN?
\item How does Dropout impact the overestimation phenomenon in DQN?
\item How does Dropout affect the quality of the learned policies?
\end{itemize}
To that end, we ran Dropout-DQN and DQN on one of the classic control environments to express the effect of Dropout on Variance and the learned policies quality. For the Overestimation phenomena, we ran Dropout-DQN and DQN on a Gridworld environment to express the effect of Dropout because in such environment the optimal value function can be computed exactly.

\begin{table*}
\centering
\caption{Variance comparison use Wilcoxon Sign-Ranked Test in CARTPOLE environment. DQN and its variants achieved scores(Rewards) are averaged over 10 independent learning trails.}
\begin{tabular}{|l|l|l|l|} 
\hline
~~~~~~~ Variance Comparison                                               & \begin{tabular}[c]{@{}l@{}}Before Dropout(DQN)\\~~~~~ Avg.~~~ (Std.)\end{tabular} & \begin{tabular}[c]{@{}l@{}}~ After Dropout\\~~~ Avg.~~~ (Std.)\end{tabular} & \begin{tabular}[c]{@{}l@{}}Wilcoxon Sign-Ranked Test \\~~~ Statistic~~~~~~ (p-value)\end{tabular}  \\ 
\hline
\begin{tabular}[c]{@{}l@{}}DQN~ VS~ Gaussian Dropout DQN \\ \end{tabular} & \begin{tabular}[c]{@{}l@{}}~ 108.020~~ (54.932)\\\end{tabular}                    & 117.871~ (46.846)                                                           & ~~~~ 175584.0~~ (3.005e-16)                                                                        \\ 
\hline
DQN~ VS~ Variational Dropout DQN                                          & \begin{tabular}[c]{@{}l@{}}~ 108.020~ (54.932)\\\end{tabular}                     & 51.490~~~ (28.075)                                                          & ~~~~ 30695.0~~ (1.256e-127)                                                                        \\
\hline
\end{tabular}
\label{tab1}
\end{table*}

\subsection{Classic Control Environment}
To evaluate the Dropout-DQN, we employ the standard reinforcement learning (RL) methodology, where the performance of the agent is assessed at the conclusion of the training epochs. Thus we ran ten consecutive learning trails and averaged them. We have evaluated Dropout-DQN algorithm on CARTPOLE problem from the Classic Control Environment. The game of CARTPOLE was selected due to its widespread use and the ease with which the DQN can achieve a steady state policy.
For the experiments, fully connected neural network architecture was used. It was composed of two hidden layers of 128 neurons and two Dropout layers between the input layer and the first hidden layer and between the two hidden layers. To minimize the
DQN loss, ADAM optimizer was used\cite{b25}.\\
We detected the variance between DQN and Dropout-DQN  visually and numerically as Figure \ref{fig3} and Table \ref{tab1} show.

The results in Figure \ref{fig3} show that using DQN with different Dropout methods result in better-preforming policies and less variability as the reduced standard deviation between the variants indicate to. In table 1, Wilcoxon Sign-Ranked test was used to analyze the effect of Variance before applying Dropout (DQN) and after applying Dropout (Dropout methods DQN). There was a statistically significant decrease in Variance (14.72\% between Gaussian Dropout and DQN, 48.89\% between Variational Dropout and DQN). Furthermore one of the Dropout methods outperformed DQN score.\\
The findings indicate that Dropout can effectively reduce the variance and overestimation issues in DQN, leading to more stable learning curves and notably enhanced performance.

\subsection{Gridworld}
The Gridworld problem (Figure \ref{fig4}) is a common RL benchmark. Its relatively small state space permits the Experience Replay (ER) buffer to store all possible state-action pairs. Moreover, this setup allows for the precise computation of the optimal action value function.
\subsubsection{ENVIRONMENT SETUP}
For this experiment, we designed a customized environment modeled after the Gridworld problem (Figure \ref{fig4}) the state space contains pairs of points from a 2D discrete grid ($ S = {(x,y)_x,_y\in0,1,2,3,4}$). The environment includes four possible actions, each corresponding to movement in one of the cardinal directions. The rewards are set up such that $r = +1 $ is awarded in the goal state G (Goal), and $r = -1 $ otherwise.

A fully connected neural network architecture was used. It was composed of two hidden layers of 128 neurons and two Dropout layers between the input layer and the first hidden layer and between the two hidden layers. ADAM optimizer for the minimization\cite{b25}.
\begin{figure}[htbp]
\centerline{\includegraphics[width=.5\textwidth]{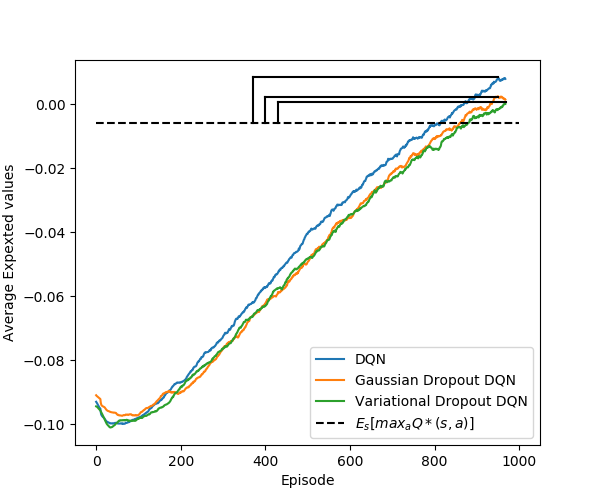}}
\caption{Average Expected value in Gridworld. Dropout methods on DQN lead to less overestimation (positive-bias). The lines are averages over 50 independent learning trials. The dotted line represents the Optimal Policy.}
\label{fig5}
\end{figure}
\subsubsection{OVERESTIMATION}
Figure \ref{fig5} demonstrates that using Dropout methods in DQN reduce the overestimation from the optimal policy. Despite that Gridworld environment is not suffering from intangible overestimation that can distort the overall cumulative rewards but reducing overestimation leads to more accurate predictions.

\subsection{The Learned Policies}
\begin{figure*}
\begin{minipage}[b]{1\linewidth}
\centering
\subfloat[]{\includegraphics[width=5cm]{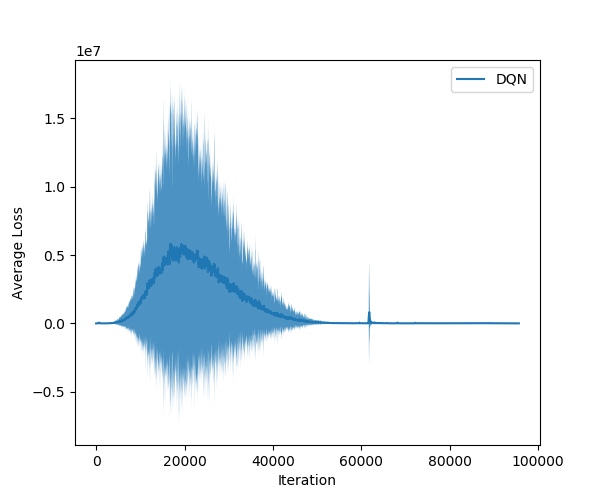}}\quad
\subfloat[]{\includegraphics[width=5cm]{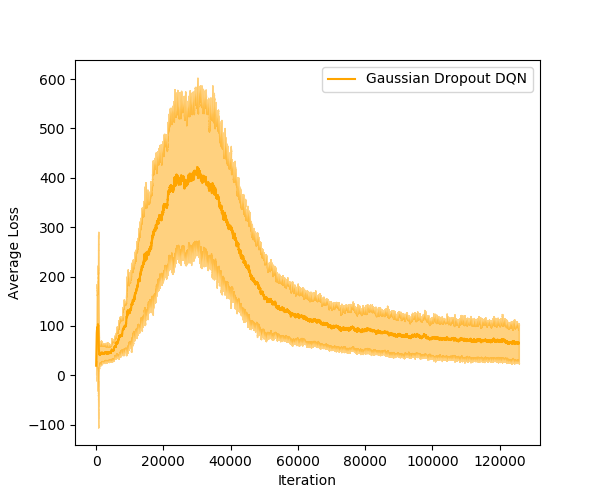}}
\subfloat[]{\includegraphics[width=5cm]{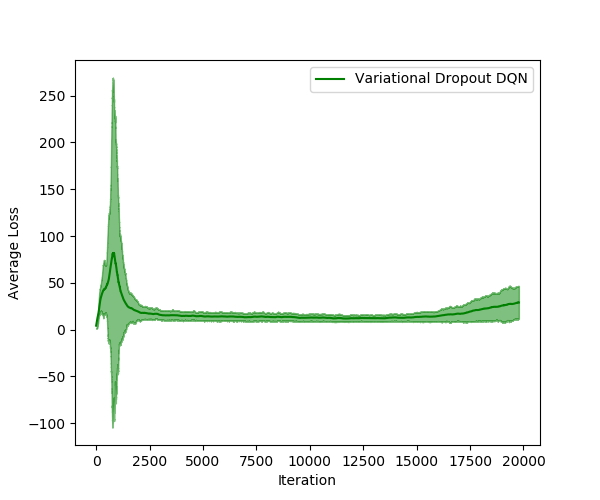}}
\caption{The Average Loss metrics on Dropout DQN and DQN in CARTPOLE environment, iterations represent minimum times of soft updated loss of 10 independent learning trails.}
\label{fig6}
\end{minipage}
\end{figure*}

Figure \ref{fig6} shows the loss metrics of the three algorithms in CARTPOLE environment, this implies that using Dropout-DQN methods introduce more accurate gradient estimation of policies through iterations of different learning trails than DQN. The rate of convergence of one of Dropout-DQN methods has done more iterations than DQN under the same assumption for both algorithms. It has been theoretically proven\cite{b26} that a large number of iterations creates a good quality policy that its corresponding Q values function converge to the optimal Q function, and now it's empirically demonstrated. Both Dropout-DQN algorithms have lower loss than DQN, this means that more accurate predictions of the value of the current policy which might not be the optimal policy but at least have a small deviation of loss between different policies and with all mentioned factors above lead to less variance in cumulative rewards and less overestimation of certain policies.
\section{Conclusion and Future Directions}

In this study, we proposed and experimentally analyzed the benefits of incorporating the Dropout technique into the DQN algorithm to stabilize training, enhance performance, and reduce variance. Our findings indicate that the Dropout-DQN method is effective in decreasing both variance and overestimation. However, our experiments were limited to simple problems and environments, utilizing small network architectures and only two Dropout methods.

Future work is going to including experimenting more Dropout methods on more challenging problems in complex Environments, e.g. Arcade Learning Environment(ALE).

Dropout-DQN represents a straightforward modification that seamlessly complements various DQN variants, including Prioritized Experience Replay\cite{b27}, Double Q Learning\cite{b10}, Dueling Q Learning\cite{b28}, Optimality Tightening\cite{b29}, and Unifying Count-Based Exploration and Intrinsic Motivation\cite{b30}. Exploring the additional benefits of integrating Dropout with these existing variants would be a valuable area of study.

\end{document}